\documentclass[pdflatex,sn-mathphys-ay, iicol]{sn-jnl}

\usepackage{graphicx}
\usepackage{multirow}
\usepackage{amsmath,amssymb,amsfonts}
\usepackage{amsthm}
\usepackage{mathrsfs}
\usepackage[title]{appendix}
\usepackage{xcolor}
\usepackage{textcomp}
\usepackage{manyfoot}
\usepackage{booktabs}
\usepackage{algorithm}
\usepackage{algorithmicx}
\usepackage{algpseudocode}
\usepackage{listings}
\theoremstyle{thmstyleone}

\theoremstyle{thmstyletwo}

\theoremstyle{thmstylethree}

\begin{document}

\title[Article Title]{Gen-Fab: A Variation-Aware Generative Model for Predicting Fabrication Variations in Nanophotonic Devices}

\author*[1]{\fnm{Rambod} \sur{Azimi}}\email{rambod.azimi@mail.mcgill.ca}

\author[2]{\fnm{Yuri} \sur{Grinberg}} 

\author[3]{\fnm{Dan-Xia} \sur{Xu}} 

\author[1]{\fnm{Odile} \sur{Liboiron-Ladouceur}} 

\affil[1]{\orgdiv{Department of Electrical and Computer Engineering}, \orgname{McGill University}, \postcode{Montreal}, \state{QC}, \country{Canada}}

\affil[2]{\orgdiv{Digital Technologies Research Center, National Research Council of Canada}, \state{Ottawa}, \state{ON}, \country{Canada}}
\affil[3]{\orgdiv{Quantum and Nanotechnologies Research Center, National Research Council of Canada}, \state{Ottawa}, \state{ON}, \country{Canada}}

\abstract{Silicon photonic devices often exhibit fabrication-induced variations such as over-etching, under-etching, and corner rounding, which can significantly alter device performance. These variations are nonuniform, influenced by feature size and shape. Accurate digital twins are needed to predict the range of possible fabricated outcomes for a given design. In this paper, we introduce Gen-Fab, a conditional generative adversarial network (cGAN) based on the Pix2Pix, to predict and model uncertainty in photonic fabrication outcomes. The proposed method takes a design layout (in GDS format) as input and produces diverse high-resolution predictions similar to scanning electron microscope (SEM) images of fabricated devices, capturing the range of process variations at the nanometer scale. To enable one-to-many mapping, we inject a latent noise vector at the model’s bottleneck. We compare Gen-Fab against three baselines: (1) a deterministic U-Net predictor, (2) an inference-time Monte Carlo Dropout U-Net, and (3) an ensemble of varied U-Nets. Evaluations on an out-of-distribution dataset of fabricated photonic test structures demonstrate that Gen-Fab outperforms all baselines in accuracy and uncertainty modeling, and an additional distribution-shift analysis confirms its strong generalization to unseen fabrication geometries. Gen-Fab achieves the highest intersection-over-union (IoU) score of 89.8\%, outperforming the deterministic U-Net (85.3\%), the MC-Dropout U-Net (83.4\%), and Varying U-Nets (85.8\%), and better aligns with the distribution of real fabrication outcomes, attaining lower Kullback–Leibler divergence and Wasserstein distance.}


\maketitle

\noindent\textbf{Note:} This manuscript is a preprint of a paper accepted for publication in 
\textit{Structural and Multidisciplinary Optimization}. 
The final version of record is available at: 
\url{https://doi.org/10.1007/s00158-026-04272-3}.

\section{Introduction}\label{sec1}
Silicon photonics is an emerging platform for integrating optical components on silicon chips, enabling compact, energy-efficient, and high-speed photonic circuits. Applications span data-center interconnects, LiDAR, quantum photonics, and biosensing. Devices are typically fabricated using Complementary Metal-Oxide-Semiconductor (CMOS)-compatible processes involving lithography, etching, and material deposition~\citep{microprocessorlight2015, siliconphotonicssensors2022, shekhar2024siliconphotonics}. Silicon photonic devices are sensitive to fabrication-induced deviations such as over-etching, under-etching, and corner rounding~\citep{fabvariation, gostimirovic2022deep, gostimirovic2023improving, azimi2025sem}, which are caused by both systematic and stochastic effects during the involved processing steps~\citep{han2019ler}. These deviations are often spatially nonuniform and can degrade device performance while complicate design optimization.

\begin{figure}[t]
\centering
\includegraphics[width=0.4\textwidth]{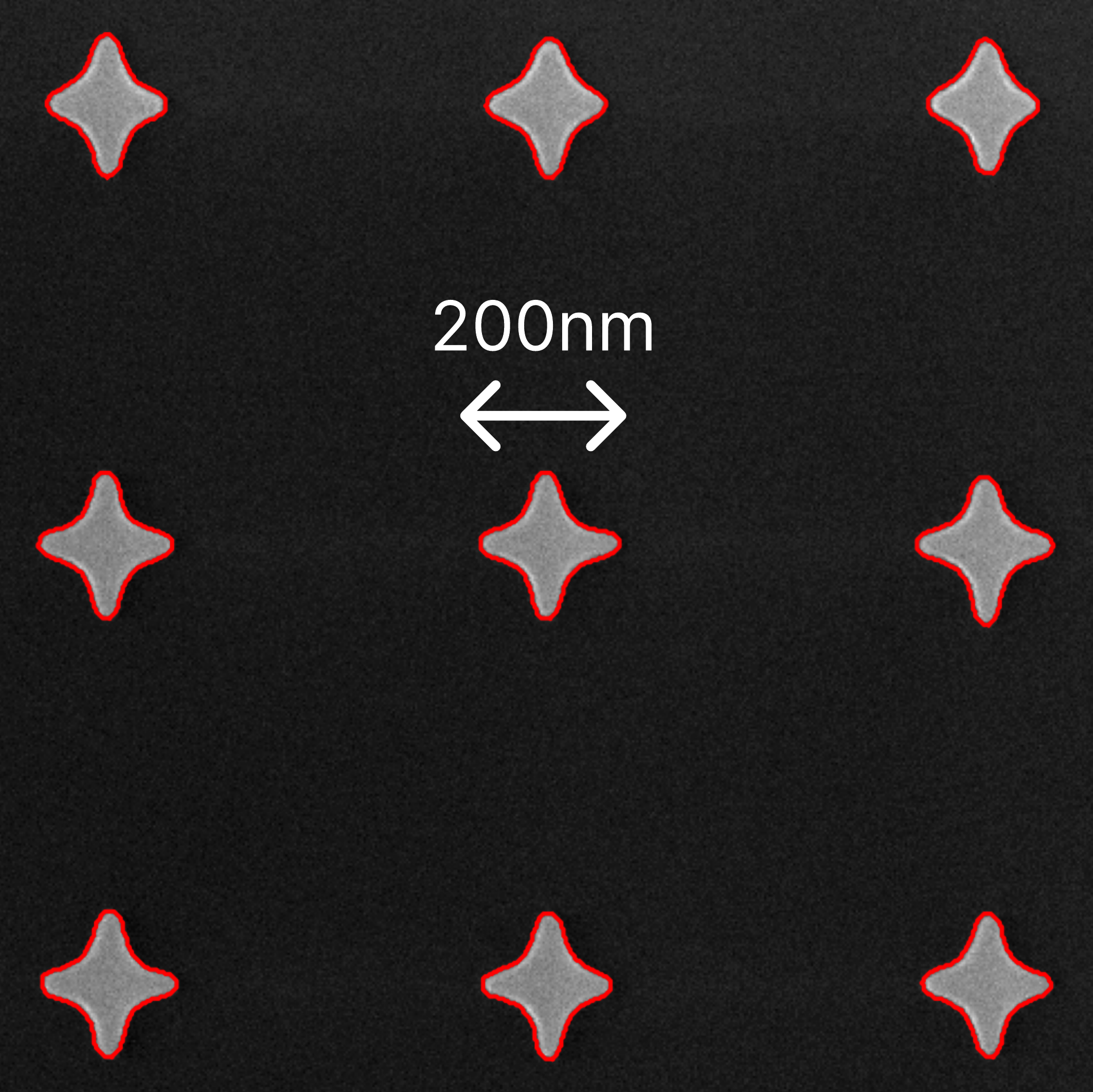}
\caption{\textbf{Overlay of SEM images from repeated fabrications of the same photonic cross design.}  
Although each fabricated structure used the same GDS file layout, visible differences emerge in edge sharpness and arm geometry, indicated in red.}\label{fig1}
\end{figure}

\begin{figure}[t]
\centering
\includegraphics[width=0.45\textwidth]{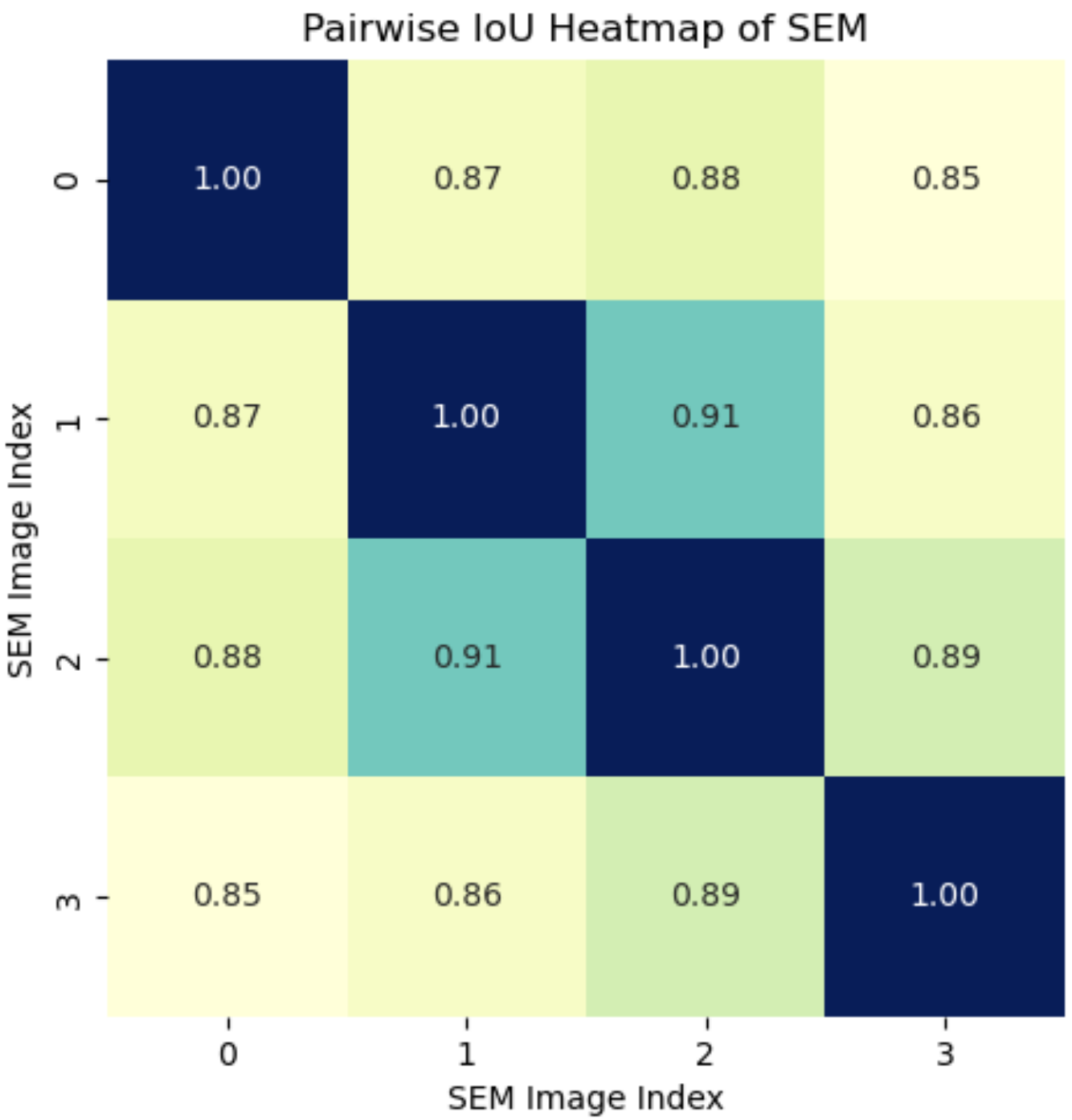}
\caption{\textbf{IoU similarity heatmap of SEM images from four fabricated crosses.}  
Each cell shows the IoU score between pairs of SEM images from four devices fabricated from the same layout.}\label{fig2}
\end{figure}

Traditional approaches such as statistical 'corner analysis' and conventional design rules are typically simplistic, lacking spatial resolution and failing to capture the complex, stochastic nature of real-world process variability~\citep{fabricationconstrainednanophotonicinversedesign, unpredictablefabvariation, james2023process}. Repeated fabrication of identical GDS layouts reveals some degree of variation in the resulting scanning electron microscope (SEM) images (see Fig.~\ref{fig1}), and quantitative measures such as Intersection-over-Union (IoU) heatmaps (Fig.~\ref{fig2}) confirm that these differences are not negligible, thereby playing a key role in determining device-level variability. As a result, there is growing interest in developing fabrication-aware digital twins that simulate a distribution of fabrication outcomes from a given layout, capturing both deterministic and stochastic aspects.

From a digital-twin perspective, such fabrication-aware modeling aims to construct a virtual, data-driven replica of the physical manufacturing process that can predict and sample realistic fabrication outcomes under uncertainty. In this context, an effective fabrication digital twin must be able to map an ideal design layout to a distribution of physically plausible manufactured realizations rather than a single deterministic prediction. This motivates the use of probabilistic, data-driven surrogate models capable of capturing both deterministic structure and stochastic process variability, marking an important stepping stone towards a full-fledged digital twin solution.

Given the structured, image-based nature of both photonic GDS layouts and their fabricated counterparts (SEM), image-to-image translation using deep learning has emerged as a compelling approach. Prior work has used convolutional neural networks (CNNs), particularly U-Net architectures~\citep{ronneberger2015unet}, to reconstruct SEM images from layout masks~\citep{gostimirovic2022deep, azimi2025sem}.
These deterministic encoder-decoder models have demonstrated high pixel-level accuracy in capturing fabrication deviations such as missing features. However, they yield a single prediction for a given input, failing to reflect the distributional spread observed in actual fabrication processes~\citep{akbari2023probabilistic}.

Among the methods proposed to address this limitation, ensemble learning has been shown to be one of the most robust and effective approaches for modeling uncertainty~\citep{modeluncertainty}, and we adopt it as a strong baseline in our experiments. To further explore uncertainty modeling without retraining multiple networks, we additionally examine an inference-time Monte Carlo (MC) Dropout U-Net baseline~\citep{dropoutdiversity}, which applies dropout layers during both training and inference to produce stochastic predictions from a single trained model. While this approach provides a Bayesian approximation of model uncertainty rather than data variability, its implementation is straightforward; therefore, it is included as another naïve alternative. Our results show that its variability does not accurately reflect the physical diversity of real fabrication outcomes. Alternative approaches such as heuristic threshold-based variability estimation, which we also experimented with, similarly fail to capture the full range of fabrication-induced variations.

In the meantime, probabilistic generative models such as variational autoencoders (VAEs)~\citep{kingma2022} and generative adversarial networks (GANs)~\citep{originalgan} have gained popularity as tools that can be used to sample high-dimensional data points from some complex unknown distribution. Among those, conditional GANs (cGANs)~\citep{mirza2014conditionalgenerativeadversarialnets}, particularly the Pix2Pix conditional GAN framework~\citep{isola2018pix2pix}, offer a principled way to model fabrication deviations in an image-to-image translation setup. While the original Pix2Pix framework was deterministic, later extensions introduced stochastic conditioning via injected latent noise vectors z~\citep{naderi2022dynamicpix2pix}, making it well-suited for capturing the one-to-many mapping in fabrication outcomes.

We chose a GAN-based conditional framework for this study. Our application involves paired high-resolution data (GDS layouts aligned to SEM images), a setting where conditional GANs such as Pix2Pix remain highly competitive due to their direct image-to-image mapping and relatively low training cost, which allows scaling of predictions to large layouts during inference. Moreover, leveraging a well-established Pix2Pix backbone enables clear comparison to prior photonic work and provides a transparent benchmark for future diffusion-based extensions. Although GANs can suffer from mode collapse or unstable training, our experiments show that careful noise injection, data augmentation, and hyperparameter tuning yield stable convergence and high-fidelity predictions in this domain. We therefore view cGAN as a practical and computationally efficient first step toward stochastic fabrication-aware modeling.

In this work, we present Gen-Fab, a variation-aware cGAN trained to predict silicon photonic fabrication outcomes. Gen-Fab builds on the Pix2Pix architecture, which in its original formulation is a deterministic pixel-to-pixel translation model with no latent noise input. In contrast, while all GANs conceptually rely on a latent variable \(z\) for sampling diversity, the standard Pix2Pix omits this and produces a single output per input. We explicitly reintroduce a latent noise vector z at the generator's bottleneck to enable stochastic generation of multiple SEM-like images from a single GDS input (we will refer to these outputs as 'generated SEMs' or simply as 'SEMs'). This placement improves structural fidelity while maintaining diversity in the outputs. From a digital-twin standpoint, Gen-Fab can serve as a core functionality of a generative digital twin of the nanophotonic fabrication process.

We benchmark Gen-Fab against three baselines: (1) a deterministic U-Net model, (2) an inference-time MC-Dropout U-Net, and (3) an ensemble of varied U-Nets (Varying U-Nets) intended to approximate fabrication variation. Our evaluation considers several distribution-level metrics, our own pixel-based metric as well as standard ones such as Kullback-Leibler divergence (KL-D) and Wasserstein distance (W-D), to assess model fidelity and uncertainty modeling~\citep{bai2019approximability, pmlrv206ting}. We conducted experiments on a curated dataset comprising GDS-SEM pairs across six types of nominal nanophotonic structures fabricated under identical conditions. To further validate model robustness, we also analyzed the feature-space distribution shift between the training and out-of-distribution (OOD) test sets, confirming that our evaluation reflects genuine generalization rather than dataset overlap. Our proposed model attained noticeably improved scores on all metrics across all presented structures and better models the observed fabrication variability based on further qualitative assessment. These results establish Gen-Fab as a practical tool for variation-aware prediction, with immediate applications in robust photonic design and digital twin development.

\section{Methodology}\label{sec3}

\begin{figure*}[t]
\centering
\includegraphics[width=0.79\textwidth]{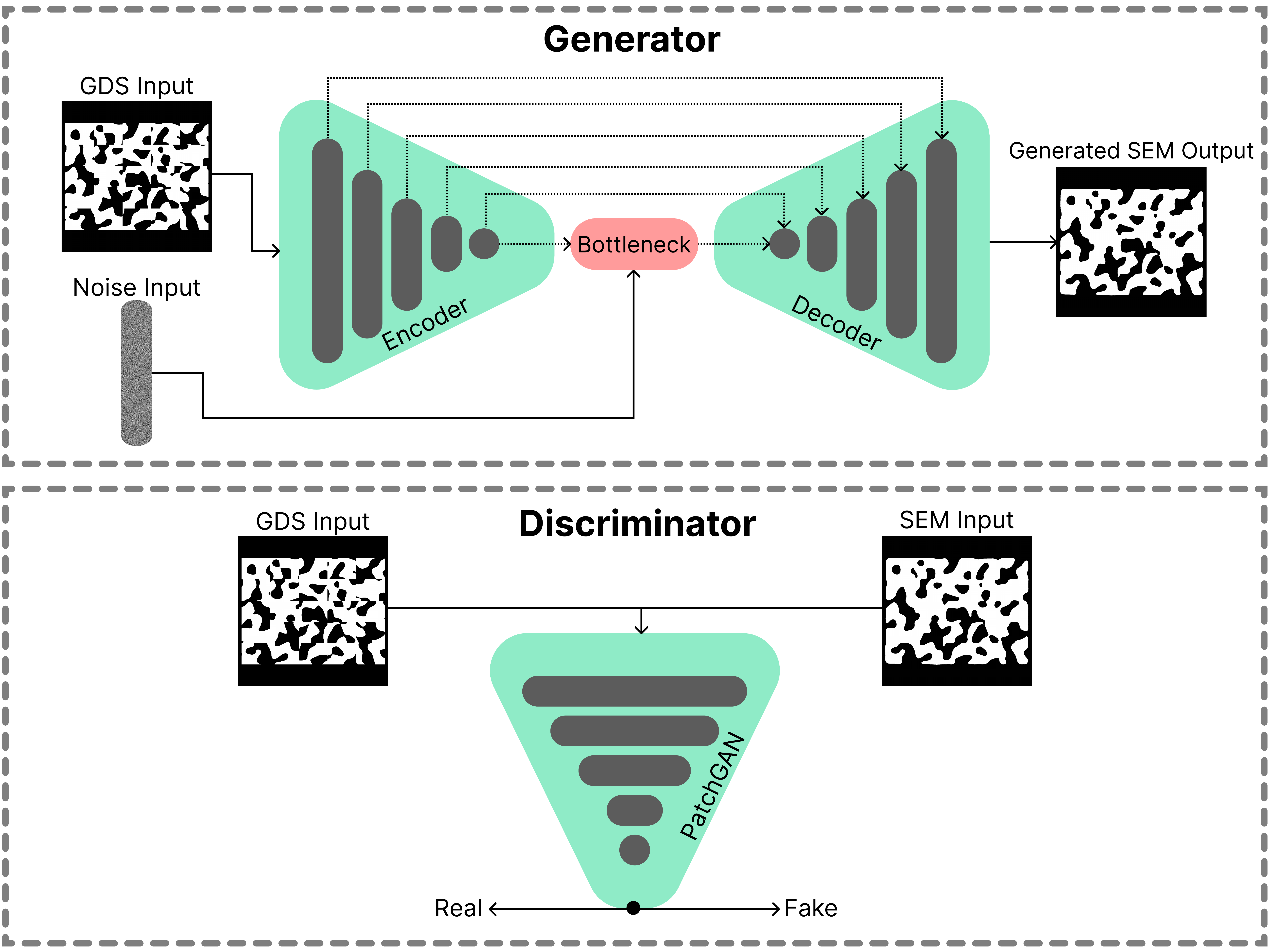}
\caption{%
        \textbf{Overview of the Gen-Fab architecture with a noise-injected generator.}
        \textbf{Top: Generator.} The model receives a GDS layout and a latent noise vector injected into the bottleneck, enabling it to produce diverse SEM outputs from the same input.
        \textbf{Bottom: Discriminator.} The PatchGAN discriminator takes the input GDS and corresponding SEM (real or generated) and evaluates $70\times70$ pixel patches to ensure outputs are locally indistinguishable from real SEMs.}\label{fig7}
\end{figure*}

\begin{figure*}[t]
\centering
\includegraphics[width=0.78\textwidth]{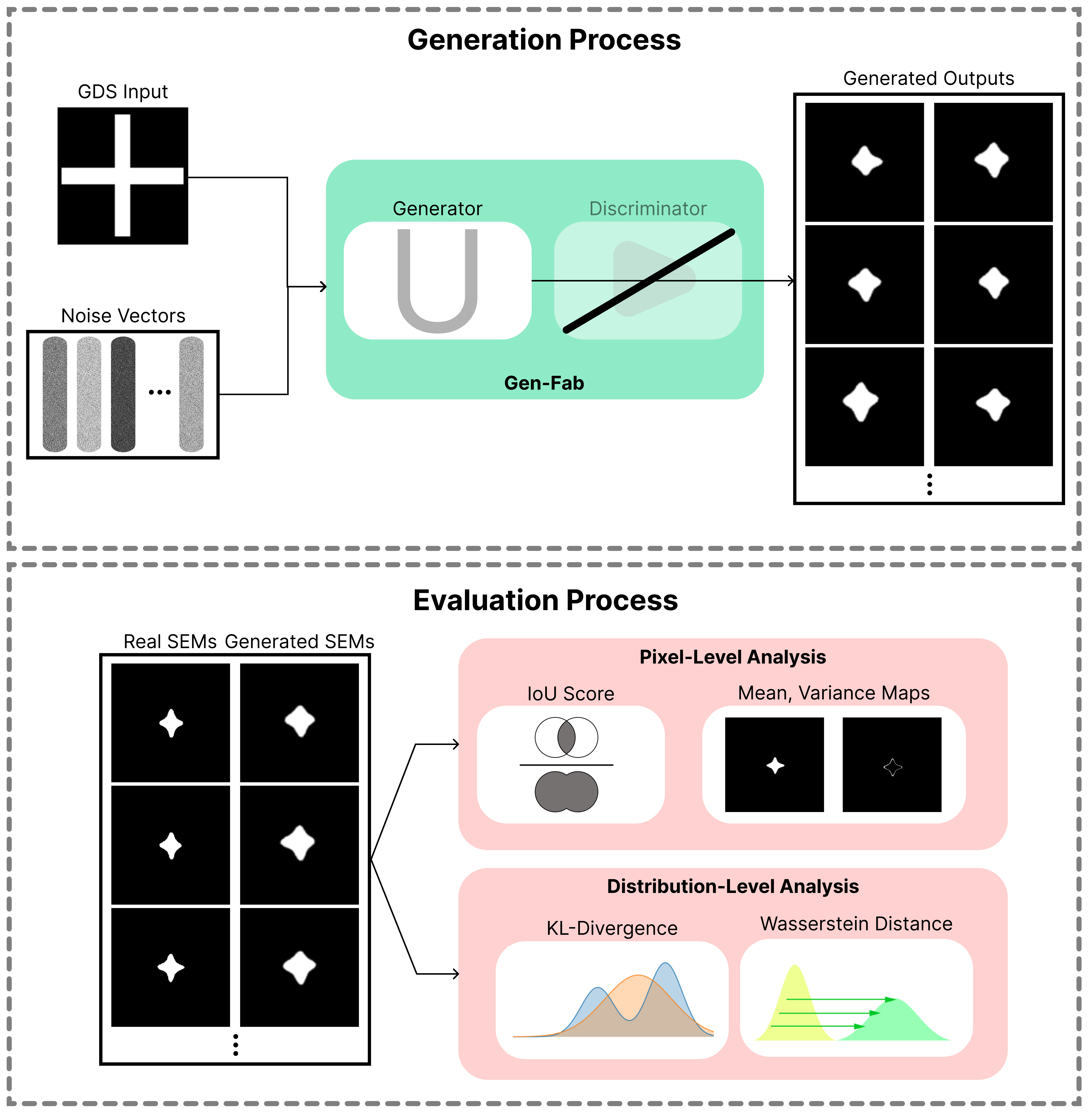}
\caption{%
        \textbf{Overview of Gen-Fab’s generation and evaluation pipeline.}
        \textbf{Top: Generation Process.} The Gen-Fab generator takes a GDS input and multiple noise vectors to produce a diverse SEM predictions, while the discriminator is discarded after training.
        \textbf{Bottom: Evaluation Process.} Generated SEMs are compared with real SEMs using pixel-level metrics (IoU, variance map) and distribution-level metrics (KL-D and W-D).
    }\label{fig8}
\end{figure*}

\subsection{Conditional Generative Modeling Approach}
We formulate fabrication variation prediction as a conditional generative modeling task. Instead of a deterministic one-to-one mapping from layout to fabricated outcome, we seek to learn a distribution of possible outcomes $P(Y|X)$ given a design $X$. To achieve this, we build upon the Pix2Pix conditional GAN framework~\citep{mirza2014conditionalgenerativeadversarialnets, isola2018pix2pix}, which couples a generator $G$ and discriminator $D$ in an adversarial training process. The generator $G$ learns to translate an input GDS layout into an output predicted SEM that is aimed to be indistinguishable from a real fabricated result, while $D$ learns to differentiate real vs. generated SEM images.

Unlike the original Pix2Pix framework, which omits a latent variable for deterministic image translation, Gen-Fab reintroduces a stochastic latent code $z$ at the bottleneck layer of the U-Net generator, enabling multiple plausible fabrication outcomes to be sampled for a given input layout.

\subsection{Gen-Fab Architecture with Noise Injection}
The proposed Gen-Fab model extends the Pix2Pix architecture to enable variation-aware prediction. Fig.~\ref{fig7} provides an overview of this architecture. The generator $G$ follows a U-Net style encoder–decoder with skip connections, mapping an input layout image $X$ to an output SEM image $\hat{Y}$. The encoder first processes the input design $X$ through a series of convolutional downsampling layers, yielding a coarse feature representation at the bottleneck. At this point, the $d$-dimensional noise $z$ is spatially expanded to match the bottleneck feature map dimensions by tiling it across height and width. This forms an augmented bottleneck feature that carries both deterministic layout-conditioned features and stochastic latent information. The decoder then processes this combined feature map, mapping it back up through a series of convolutional upsampling layers to produce an image output. As a result, Gen-Fab generates a distribution of outputs $\hat{Y}$ for a single input $X$ by sampling different $z$, effectively capturing real-world fabrication variations.

The discriminator $D$ uses the PatchGAN architecture from Pix2Pix~\citep{demir2018patchbased}. Instead of evaluating the entire image, it processes small $70\times70$ pixel patches from the input–output pair and judges whether each patch resembles a realistic fabrication result. The overall Gen-Fab architecture thus consists of: (1) a U-Net generator $G(X,z)$ producing diverse SEM-like predictions, and (2) a conditional PatchGAN discriminator $D$ distinguishing real from fake outcomes.

\subsection{Training Objectives}
Training Gen-Fab combines of adversarial and reconstruction objectives to balance realism and fidelity. Let $X$ be the input layout, $Y$ the corresponding true SEM image from fabrication, and $z$ the latent noise vector. The discriminator maximizes the log-likelihood of correctly classifying real versus fake pairs, following the conditional GAN objective~\citep{isola2018pix2pix}:
\begin{multline}\label{eq:adv_D}
L_{D} = -\mathbb{E}_{X,Y}[\log D(X, Y)] \\
\quad -\; \mathbb{E}_{X,z}[\log(1 - D(X, G(X,z)))]~,
\end{multline}
where $D(X,Y)$ is the discriminator’s estimated probability that $(X,Y)$ is a real layout-SEM pair. Meanwhile, the generator’s adversarial loss $L_{\text{GAN}}$ can be written as
\begin{equation}\label{eq:adv_G}
L_{\text{GAN}}(G) = -\mathbb{E}_{X,z}[\log D(X, G(X,z))]~,
\end{equation}
which encourages $G$ to generate outputs that $D$ will judge as real. In addition, we include an $L_1$ reconstruction loss written as
\begin{equation}
\mathcal{L}_{\text{L1}}(G) = \mathbb{E}_{X, Y, z} \left[ \left\| Y - G(X, z) \right\|_1 \right]
\end{equation}
that penalizes pixel-level errors between the generated image and the ground truth. We weight this term with a hyperparameter $\lambda$ to balance sharpness vs. fidelity. Following the original Pix2Pix implementation~\citep{isola2018pix2pix}, we set the weighting parameter $\lambda = 100$ to emphasize reconstruction fidelity while still allowing the adversarial loss to sharpen details. This value has been widely adopted in image-to-image translation tasks because it balances the adversarial and pixel-level losses, producing outputs that are both structurally accurate and visually realistic. In our preliminary experiments, smaller $\lambda$ values (10 or 50) led to blurrier reconstructions, while larger values suppressed stochastic diversity.
The total objective for the generator is thus a weighted sum of adversarial and reconstruction terms:
\begin{equation}\label{eq:total_G}
\mathcal{L}_{G} = \mathcal{L}_{\text{GAN}}(G) + \lambda \, \mathcal{L}_{\text{L1}}(G)
\end{equation}
which $G$ attempts to minimize. This two-player minimax game is solved via alternating optimization of $G$ and $D$. The adversarial component $L_{\text{GAN}}$ guides $G$ to match the distribution of real SEM images, while the $L_1$ term ensures that the main device features in $\hat{Y}=G(X,z)$ align with those in the true SEM $Y$. By injecting random $z$ during training, the generator learns that it must produce an output that not only fools $D$ and matches $Y$ (for the current $z$) but also that it has the flexibility to yield different valid $Y$ for different $z$.

\subsection{Training Procedure}
We train Gen-Fab using the Adam optimizer~\citep{kingma2017adammethodstochasticoptimization}, a stochastic gradient–based method with mini-batch updates. Algorithm~\ref{alg:training} outlines the training procedure. At each iteration, a batch of paired samples ${(X_i, Y_i)}$ is fetched from the training set, and a random latent vector $z_i$ is sampled for each. The generator produces $\hat{Y}_i = G(X_i, z_i)$, a fake SEM image for each input layout. The discriminator is then updated by comparing real pairs $(X_i, Y_i)$ with fake pairs $(X_i, \hat{Y}_i)$. We compute the discriminator loss $L_D$ and take a gradient step on $D$’s parameters $\theta_D$ to improve real–fake discrimination. Next, the generator is updated: we compute the generator’s loss $L_G$, which includes the adversarial term and the $L_1$ term. After sufficient training steps, this procedure yields a generator that can produce a range of realistic outputs for each input layout.

\begin{algorithm}[t]
\caption{Gen-Fab Training Procedure}
\label{alg:training}
\begin{algorithmic}
\Require Paired data $\{(X_i, Y_i)\}_{i=1}^N$, noise dim $d$, weights $\lambda_{\text{GAN}}{=}1$, $\lambda_{\text{L1}}{=}100$, iterations $T$
\State Initialize generator parameters $\theta_G$, discriminator parameters $\theta_D$
\For{$t = 1$ to $T$}
  \State Sample $m$ pairs $\{(X_j, Y_j)\}$ from training set
  \State Sample $z_j \sim \mathcal{N}(0, I)$ for $j = 1, \ldots, m$
  \State Generate $\hat{Y}_j = G(X_j, z_j)$

  \State Compute $L_D = -\mathbb{E}[\log D(X, Y)] - \mathbb{E}[\log(1 - D(X, G(X,z)))]$
  \State Update $\theta_D$ using Adam optimizer: 
  \[
    \theta_D \leftarrow \theta_D - \eta_D \nabla_{\theta_D} L_D
  \]

  \State Compute $\mathcal{L}_{\text{GAN}} = -\mathbb{E}[\log D(X, G(X,z))]$
  \State Compute $\mathcal{L}_{\text{L1}} = \mathbb{E}[\|Y - G(X, z)\|_1]$
  \State Compute $\mathcal{L}_G = \mathcal{L}_{\text{GAN}} + \lambda_{\text{L1}} \mathcal{L}_{\text{L1}}$
  \State Update $\theta_G$ using Adam optimizer: 
  \[
    \theta_G \leftarrow \theta_G - \eta_G \nabla_{\theta_G} \mathcal{L}_G
  \]
\EndFor
\State \textbf{Note:} Parameter updates are written in simplified gradient-descent form for clarity; in practice, Adam~\citep{kingma2017adammethodstochasticoptimization} with default moment estimates and bias correction is used. \end{algorithmic} \end{algorithm}

During training, the noise vector $z$ is resampled on every iteration for every training example. This means the generator is trained across many different $z$ values over the course of training, preventing it from simply learning a fixed deterministic mapping. This training process allows the generator to utilize the noise input to produce subtly different outcomes, reflecting the fabrication-induced variability, while the overall device shape remains consistent with the input layout.

\subsection{Implementation and Training Details}
We implemented Gen-Fab using the PyTorch deep learning framework~\citep{pytorch}. The U-Net generator architecture has eight downsampling/upsampling layers with skip connections at corresponding resolutions. Instance normalization~\citep{ulyanov2017instancenormalizationmissingingredient} is applied to convolutional layers in both $G$ and $D$, which helped training converge on our relatively small dataset. The PatchGAN discriminator has five convolutional layers with increasing feature counts (64, 128, 256, 512, 512) and a final $1\times1$ convolution to produce the output map; it uses leaky ReLU activations~\citep{xu2015empiricalevaluationrectifiedactivations} and no normalization in order to preserve signal diversity. For optimization, we used the Adam optimizer~\citep{kingma2017adammethodstochasticoptimization} for both $G$ and $D$ with learning rate $,\eta_G = \eta_D = 2\times10^{-4}$. Following common GAN practice, we set momentum terms $\beta_1 = 0.5$, $\beta_2 = 0.999$ to stabilize training. The latent noise dimension $d$ is a tunable hyperparameter. We set $d=16$ based on preliminary experiments, balancing diversity and stability. Larger $d$ led to unstable training, while smaller $d$ limited stochastic expressiveness. All models were trained on $2048 \times 2048$ pixel patches with a mini-batch size of 4. This batch size was selected to balance GPU memory constraints with stable gradient estimates; preliminary tests with smaller batches showed no measurable improvement in convergence or generalization. All models were trained on a single NVIDIA RTX 4090 GPU (24 GB VRAM), with an average end-to-end training time of approximately 50 minutes per configuration.

\subsection{Inference (Generation) Process}
Once trained, the Gen-Fab generator is used to produce multiple outputs for each new input layout. The discriminator is discarded at inference. To generate a set of $M$ possible fabricated outcomes for a given design $X$, we simply sample $M$ independent noise vectors ${z^{(1)}, z^{(2)}, \dots, z^{(M)}}$ from the latent distribution and feed them through the generator: $\hat{Y}^{(m)} = G(X, z^{(m)})$ for $m=1,\ldots,M$. This yields $M$ SEM-like predictions for the single input layout. The multiple outputs can be analyzed to quantify uncertainty and compared against actual fabricated samples. The full generation pipeline is illustrated in Fig.~\ref{fig8}.

\section{Experimental Results and Analysis}\label{sec4}

\subsection{Experiment Setup}
  \subsubsection{Train Dataset}
  
\begin{figure}[t]
\centering
\includegraphics[width=0.49\textwidth]{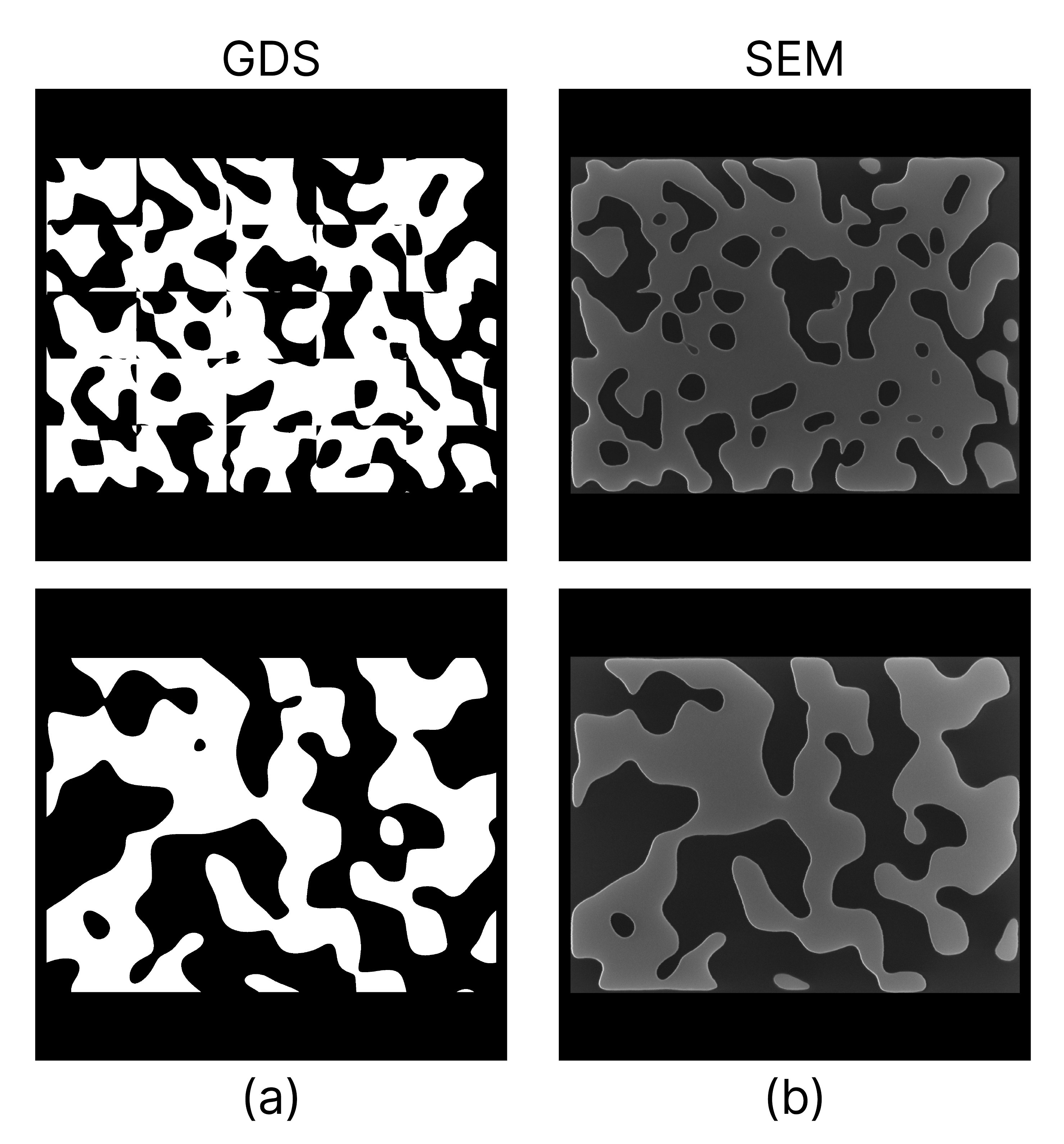}
\caption{%
    \textbf{Paired examples of GDS layouts and their fabricated SEM used for training.} 
    (a) Input GDS design (ground-truth). 
    (b) SEM of the fabricated structures, showing deviations due to the fabrication process.
    }\label{fig6}
\end{figure}

  We utilize the ANT-NanoSOI dataset, developed in collaboration with Applied Nanotools Inc. (ANT)\footnote{\url{https://www.appliednt.com/nanosoi-fabrication-service/}}, to train our Gen-Fab model. The patterns were fabricated on a 220\,nm-thick silicon-on-insulator (SOI) platform using electron-beam lithography through a multi-project wafer service. The dataset consists of non-functional patterns designed to mimic the structural characteristics of real photonic devices, in particular including free-form inverse designed structures~\citep{molesky2018}. Building on the approach of Gostimirovic et al.~\citep{gostimirovic2022deep}, we generate these random patterns using a Fourier-based procedural approach, where the underlying spatial frequency components are sampled from uniform and Gaussian distributions to emulate diverse geometric textures and spatial variations. By controlling parameters such as feature width, bend radius, and spatial frequency in the frequency domain, we can synthesize a wide range of randomized yet physically plausible patterns. Additional spatial-domain operations, including low-pass and band-pass filtering, are applied to vary feature sizes and curvatures, while random perturbations such as edge roughness, corner rounding, and placement jitter further introduce process-like variability. This method enables the generation of complex, non-repetitive geometries resembling the stochastic variability seen in fabricated nanophotonic structures, without being tied to specific device types such as Y-branches or photonic crystals. The resulting dataset provides diverse geometric motifs that allow the model to learn how fabrication imperfections manifested across varying spatial frequencies and structural complexities. Following fabrication, scanning electron microscopy (SEM) images were captured at a resolution of 1\,nm/pixel. Each GDS design layout was aligned with its corresponding SEM image to form high-quality training pairs. Each image is $2048 \times 2048$ pixels in size.

  The original dataset consists of 31 paired GDS–SEM images. To improve model performance and training stability, several pre-processing steps were applied. First, data augmentation was performed by rotating each image by $90^\circ$, $180^\circ$, and $270^\circ$, increasing the dataset size to 124 pairs. Unlike conventional patch-based methods, no downsampling or cropping was applied. Hence, each image was used at its full resolution of $2048 \times 2048$ pixels to preserve fine structural and edge-level details critical for photonic device analysis. Gen-Fab was trained directly on these full-resolution image pairs using a small batch size and mixed-precision training to efficiently manage GPU memory while maintaining spatial fidelity. The discriminator followed the standard $70 \times 70$ PatchGAN configuration~\citep{isola2018pix2pix} to evaluate the realism of local image patches across the entire image. This patch-level discrimination enforces high-frequency consistency without constraining the global image size, allowing the generator to model both local fabrication variations and global structural integrity. Prior to training, the dataset was randomly shuffled to prevent any ordering bias introduced during image collection or augmentation. Representative examples of the aligned GDS layouts and their corresponding fabricated SEM images are shown in Fig.~\ref{fig6}.

\subsubsection{Evaluation Dataset}

\begin{figure}[t]
\centering
\includegraphics[width=0.4\textwidth]{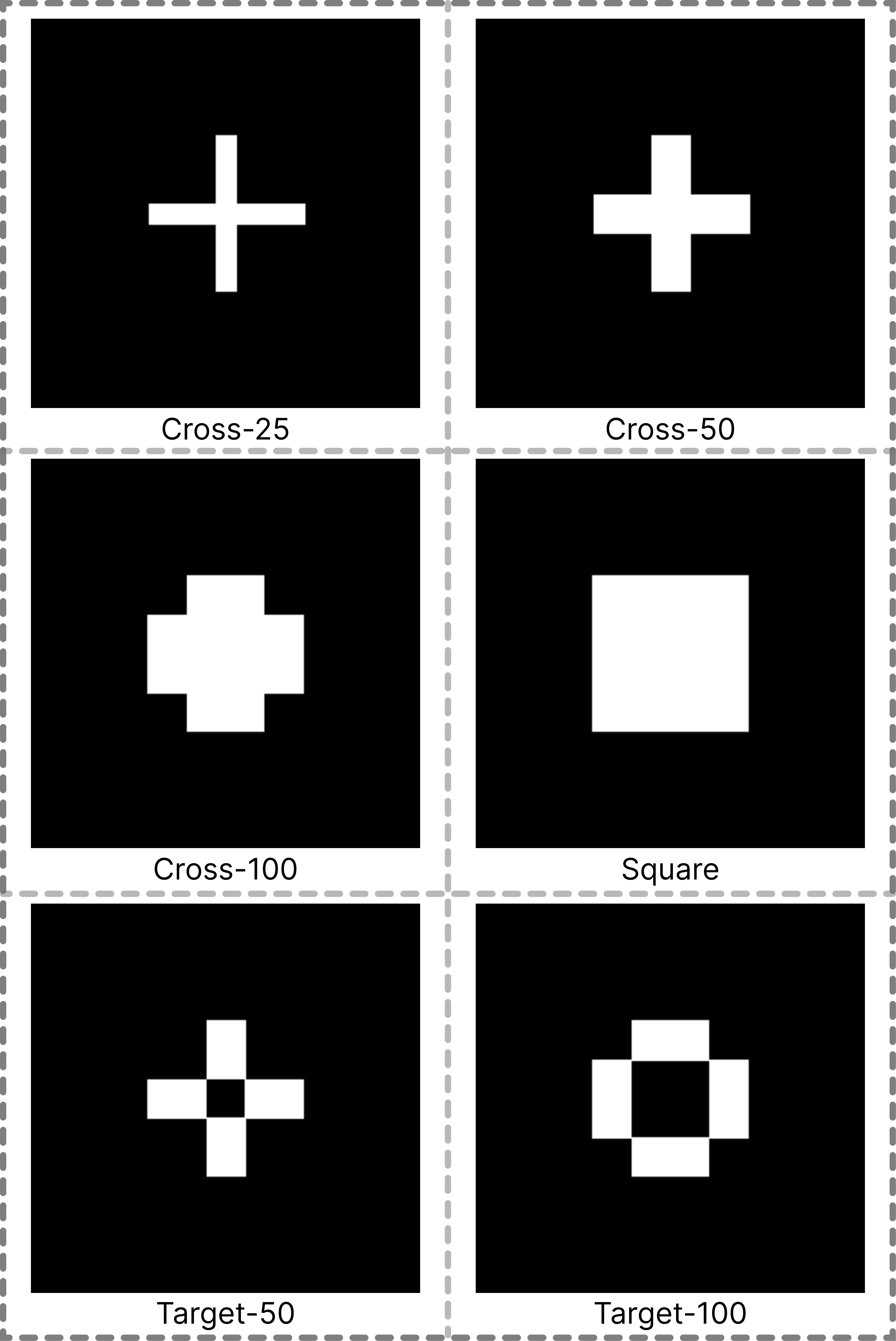}
\caption{
\textbf{Evaluation Structures Used for Model Testing.} 
We evaluate our model on six nominal GDS layouts, unseen during training. These include three cross-shaped designs, a square, and two target-like structures. The numeric suffix (25, 50, 100) indicates the arms width in nanometers. Each layout occupies a $200\text{nm} \times 200\text{nm}$ region.}\label{fig3}
\end{figure}

  To evaluate our model, we use a distinct set of six nominal test structures that were neither included during training nor drawn from the training distribution. These structures consist of Cross-25, Cross-50, Cross-100, Square, Target-50, and Target-100. Each test structure is a $200\,\text{nm} \times 200\,\text{nm}$ design composed of geometric primitives that vary in size and layout complexity. The choice of these structures serves two purposes simultaneously, first is demonstrating generalization beyond the training distribution, and second to allow easier qualitative model assessment.  

  The Cross structures vary by arm width: Cross-25 has arms $25\,\text{nm}$ wide, Cross-50 has $50\,\text{nm}$ wide arms, and Cross-100 has $100\,\text{nm}$ wide arms. The Target structures are similarly defined by cross arm width of $50\,\text{nm}$ and $100\,\text{nm}$, respectively. Each test structure includes 35 repeated fabricated SEM images. All six structures used for evaluation are shown in Fig.~\ref{fig3}.
  
  \subsubsection{Baseline Models}
  To benchmark our proposed approach, we compare Gen-Fab against three baseline models: a standard U-Net~\citep{ronneberger2015unet}, an inference-time Monte Carlo (MC) Dropout U-Net~\citep{dropoutdiversity}, and an ensemble of varied U-Nets~\citep{khoong2020busunetensembleunetframework}.

  \textbf{U-Net.} The conventional U-Net serves as a deterministic baseline model, trained on the same augmented dataset using paired GDS–SEM images~\citep{gostimirovic2022deep, azimi2025sem}. It follows the standard encoder–decoder architecture with skip connections. Once trained, the U-Net produces a single output per input, without accounting for fabrication variability.

  \textbf{MC-Dropout U-Net.} To incorporate uncertainty without retraining multiple networks, we extend the deterministic U-Net by enabling dropout layers during both training and inference following the Monte Carlo (MC) Dropout approach~\citep{dropoutdiversity}. During testing, the model is evaluated with dropout active, introducing stochastic feature masking that yields a different prediction at each forward pass. For each input layout, we perform 35 stochastic forward passes, resulting in 35 distinct predictions per sample. This technique is used to approximate Bayesian model uncertainty while keeping the overall architecture and training procedure identical to the deterministic U-Net. The number of stochastic forward passes was chosen to match the number of available fabricated SEM instances per test structure, enabling a fair distribution-level comparison. The optimal dropout rate was determined through a grid search over multiple values, with 0.1 yielding the best balance between output diversity and prediction stability.

  \textbf{Varying U-Nets.} To mimic model variability, we also evaluate an ensemble of varied U-Net models trained independently with different random initializations and data shuffling. Each independent model is used to produce an independent prediction, representing a different fabrication instance. In our experiments, the ensemble consists of 35 independently trained U-Net models. This number was chosen to match the dataset structure: for each GDS layout in the evaluation set we have 35 fabricated SEM images. Training 35 separate U-Nets with identical architecture but different random seeds allows us to generate 35 distinct predictions per layout, enabling a direct one-to-one comparison with the 35 real SEMs for each structure. This setup provides a realistic upper bound for the diversity achievable with deterministic models.

\subsection{Evaluation Metrics}
We evaluate our model using different metrics that compare distributions. First, we design our own pixel-level based metric, which captures how well individual predictions match ground-truth outcomes across two sets of images. Second, we compare sets of images using proper metrics that measure distance between distributions. For deterministic baselines like a single U-Net, which produce only one output per input, we aggregate predictions across a test set to form an empirical distribution, allowing for a fair comparison. All metrics reflect the model’s ability to reproduce the fabrication variability observed across multiple fabricated instances of different structures.

\textbf{Intersection over Union (IoU).} IoU, expressed as a percentage (\%), quantifies the pixel-wise overlap between the predicted SEM output and the ground-truth SEM image. It is computed as shown in Eq.~\eqref{eq:IoU} with each pixel treated as a binary variable; hence higher IoU values indicate that the model closely reconstructs the true geometry of the fabricated structure.

\begin{equation}\label{eq:IoU}
\text{IoU} = \frac{|A \cap B|}{|A \cup B|} = \frac{|A \cap B|}{|A| + |B| - |A \cap B|}
\end{equation}

While the IoU metric is applied to two individual images, we extend it to comparing two sets of images using pair-wise comparisons, each image taken from a respective set. For that purpose, two matching strategies are proposed and evaluated. Random matching computes the IoU between a randomly selected prediction (first set) and the ground-truth SEM (second set), simulating a naïve sample from the model's distribution. In contrast, Greedy matching selects the best match among multiple predictions by choosing the one with the highest IoU to the ground-truth.
  
\textbf{Kullback–Leibler Divergence (KL-D).} KL-D measures how one probability distribution diverges from a reference distribution. In our case, we compute \( D_{\mathrm{KL}}({\text{real}} \parallel {\text{model}}) \), evaluating how well the distribution of SEM-like images generated by the model approximates the true distribution of real SEM images of fabricated devices. Formally, given two discrete probability distributions $P$ and $Q$, the KL-D is computed as:

\begin{equation}
\text{KL}(P \| Q) = \sum_{i} P(i) \log \left( \frac{P(i)}{Q(i)} \right),
\end{equation}

where the sum runs over all pixel intensity bins. To compute KL-D, we treat each set of images as a probability distribution by first averaging them pixel-wise across the set, which results in a grayscale image where each pixel reflects the probability of being active rather than a binary value. This averaged image is then converted into a normalized histogram of pixel intensities, forming a discrete probability distribution over grayscale values. The KL-D is then computed between these histograms.

\textbf{Wasserstein Distance (W-D).} Also known as Earth Mover’s Distance (EMD), W-D quantifies dissimilarity between two probability distributions by measuring the minimum work needed to transform one distribution into the other.

Formally, for two discrete probability distributions $P$ and $Q$ over a metric space with distance function $d(x, y)$, the first-order W-D is defined as:

\begin{equation}
W(P, Q) = \inf_{\gamma \in \Pi(P, Q)} \mathbb{E}_{(x, y) \sim \gamma} [d(x, y)],
\end{equation}

where $\Pi(P, Q)$ is the set of all joint distributions $\gamma(x, y)$ whose marginals are $P$ and $Q$. Similarly to the KL-D, we first convert each set of SEM images (real or generated) into a normalized histogram of pixel intensities, treating each histogram as a discrete probability distribution. Then, for each nominal structure, we calculate the pairwise W-D between every generated and real histogram, and report the average as the final metric. As with KL-D, lower values indicate better distributional alignment.

\subsection{Quantitative Results}

\begin{table*}[t]
\caption{IoU-based distribution scores (\%) for different model types and matching strategies across test structures}\label{tab:iou_scores}
\begin{tabular*}{\textwidth}{@{\extracolsep\fill}llccccccc}
\toprule
Model Type & Matching Type & C25 & C50 & C100 & Square & T50 & T100 & Average IoU \\
\midrule
U-Net & - & 75.6 & 81.3 & 93.3 & 92.7 & 83.4 & 85.5 & 85.3 \\
MC-Dropout U-Net & Random & 72.8 & 81.0 & 84.6 & 87.3 & 83.0 & 75.3 & 80.7 \\
Varying U-Nets & Random & 75.9 & 82.9 & 92.0 & 92.4 & 85.0 & 83.1 & 85.2 \\
Gen-Fab & Random & 78.2 & 87.3 & 94.3 & 94.4 & 90.4 & 87.6 & 88.7 \\
MC-Dropout U-Net & Greedy & 76.3 & 84.4 & 85.9 & 90.0 & 85.1 & 78.7 & 83.4 \\
Varying U-Nets & Greedy & 77.1 & 83.2 & 92.3 & 92.7 & 85.5 & 84.0 & 85.8 \\
Gen-Fab & Greedy & \textbf{80.2} & \textbf{87.7} & \textbf{94.8} & \textbf{96.2} & \textbf{90.6} & \textbf{89.1} & \textbf{89.8} \\
\botrule
\end{tabular*}
\footnotetext{The Gen-Fab model with Greedy matching consistently outperforms all other baselines across all structural categories, achieving the highest overall IoU of 89.8\%.}
\end{table*}

  \subsubsection{IoU-Based Distribution Comparison}
  Table~\ref{tab:iou_scores} summarizes the IoU scores (\%) of models across the evaluation set, along with their averages. It can be observed that the IoU values are correlated with the structure size and complexity. Smaller features (e.g. Cross-25, Target-50) are reproduced with lower fidelity as compared to the larger or simpler features (Cross-100, Square), as expected in fabrication. For all these cases, the Gen-Fab model consistently achieves the highest IoU across all structure types. In particular, the Gen-Fab model with Greedy matching attains the best performance, achieving an overall IoU of 89.8\%, outperforming the next best method, Varying U-Nets with Greedy matching (85.8\%). This advantage is consistently reflected across individual categories. The MC-Dropout U-Net baseline shows lower overall IoU values of 80.7\% (Random) and 83.4\% (Greedy), below both the deterministic U-Net (85.3\%) and Varying U-Nets (85.2\% Random, 85.8\% Greedy), suggesting that dropout-based stochasticity fails to capture structural variability effectively.

  Notably, even under the Random matching, Gen-Fab achieves an IoU of 88.7\%, outperforming the deterministic U-Net (85.3\%), the MC-Dropout U-Net (80.7\% Random, 83.4\% Greedy), and the Varying U-Nets (85.2\% Random, 85.8\% Greedy).
  
  \subsubsection{Distributional Fidelity (KL-D and W-D)}
  
\begin{table*}[t]
  \caption{Comparison of KL-D and W-D between predicted and real SEM distributions across different model types}\label{tab:kl_wd_combined}
  \begin{tabular*}{\textwidth}{@{\extracolsep\fill}lcccccc}
  \toprule
  \multirow{2}{*}{Structure} & \multicolumn{3}{c}{KL-Divergence $\downarrow$} & \multicolumn{3}{c}{Wasserstein Distance $\downarrow$} \\
  \cmidrule(lr){2-4} \cmidrule(lr){5-7}
    & MC U-Net & Varying U-Nets & Gen-Fab & MC U-Net & Varying U-Nets & Gen-Fab \\
  \midrule
Cross-25        & 2.5754 & 0.9471 & \textbf{0.2657} & 0.3010 & 0.1831 & \textbf{0.1693} \\
Cross-50        & 1.6990 & 0.8150 & \textbf{0.6949} & 0.1803 & 0.1498 & \textbf{0.0980} \\
Cross-100       & 1.4570 & 0.1939 & \textbf{0.1859} & 0.1501 & 0.0692 & \textbf{0.0536} \\
Square          & 1.2188 & 0.2711 & \textbf{0.1277} & 0.1252 & 0.0673 & \textbf{0.0542} \\
Target-50       & 1.4543 & 0.5339 & \textbf{0.2335} & 0.1633 & 0.1281 & \textbf{0.1249} \\
Target-100      & 2.9397 & 0.4230 & \textbf{0.2982} & 0.2872 & 0.1322 & \textbf{0.1172} \\
\botrule
\end{tabular*}
\footnotetext{Gen-Fab achieves the lowest KL-D and W-D in nearly all categories, confirming its ability to replicate not only the expected structure but also the stochastic nature of real SEM outputs with higher fidelity than deterministic, MC-Dropout, and Varying U-Nets baselines.}
\end{table*}

  We evaluate the alignment between predicted outputs and real SEM image distributions using two statistical metrics, KL-D and W-D, as shown in Table~\ref{tab:kl_wd_combined}. Lower values indicate better alignment with the true distribution of fabricated outcomes.

  The results reveal a similar trend to the IoU-based metric: the Gen-Fab model consistently achieves the lowest KL-D and W-D across all structure types, often by a large margin. For example, in the Cross-25, Gen-Fab achieves a KL-D of 0.2657 compared to 0.9471 for the Varying U-Nets. Similar reductions are observed for Cross-50 (0.6949 vs. 0.8150) and Target-50 (0.2335 vs. 0.5339). In contrast, the MC-Dropout U-Net exhibits substantially higher divergence values across all structures (e.g., KL-D of 2.5754 on Cross-25 and 2.9397 on Target-100), indicating that dropout-induced stochasticity does not align with the true fabrication distribution, which is expected since dropout models epistemic (model) uncertainty rather than aleatoric (data) uncertainty~\citep{dropoutdiversity}.

  The same trend holds for W-D: Gen-Fab achieves the lowest W-D in every case (0.1693 on Cross-25 vs. 0.1831 for Varying U-Nets).

  \subsubsection{Hyperparameter Study of Gen-Fab Configurations}
  
  We train Gen-Fab using multiple configurations that differ in three main aspects: data augmentation, latent space dimensionality (16–128), and number of training steps (5k–20k).

  Notably, 9 out of the 10 best models employed data augmentation, demonstrating that exposing the model to augmented structures enhances its robustness. The highest-performing model (88.7\% IoU) used augmentation, a latent dimension of 16, and 10k training steps.

  Overall, moderate latent dimensions (16–64) with augmentation yield consistently high IoUs, especially with sufficient training. Larger latent dimensions do not outperform smaller ones, suggesting diminishing returns beyond a certain size.

  These results indicate that controlled stochasticity, introduced through data augmentation and a well-sized latent space, is a key factor in Gen-Fab’s success. Additionally, sufficient training duration remains essential for convergence and performance stability.
  
\subsection{Qualitative Results}
\begin{figure}[t]
\centering
\includegraphics[width=0.49\textwidth]{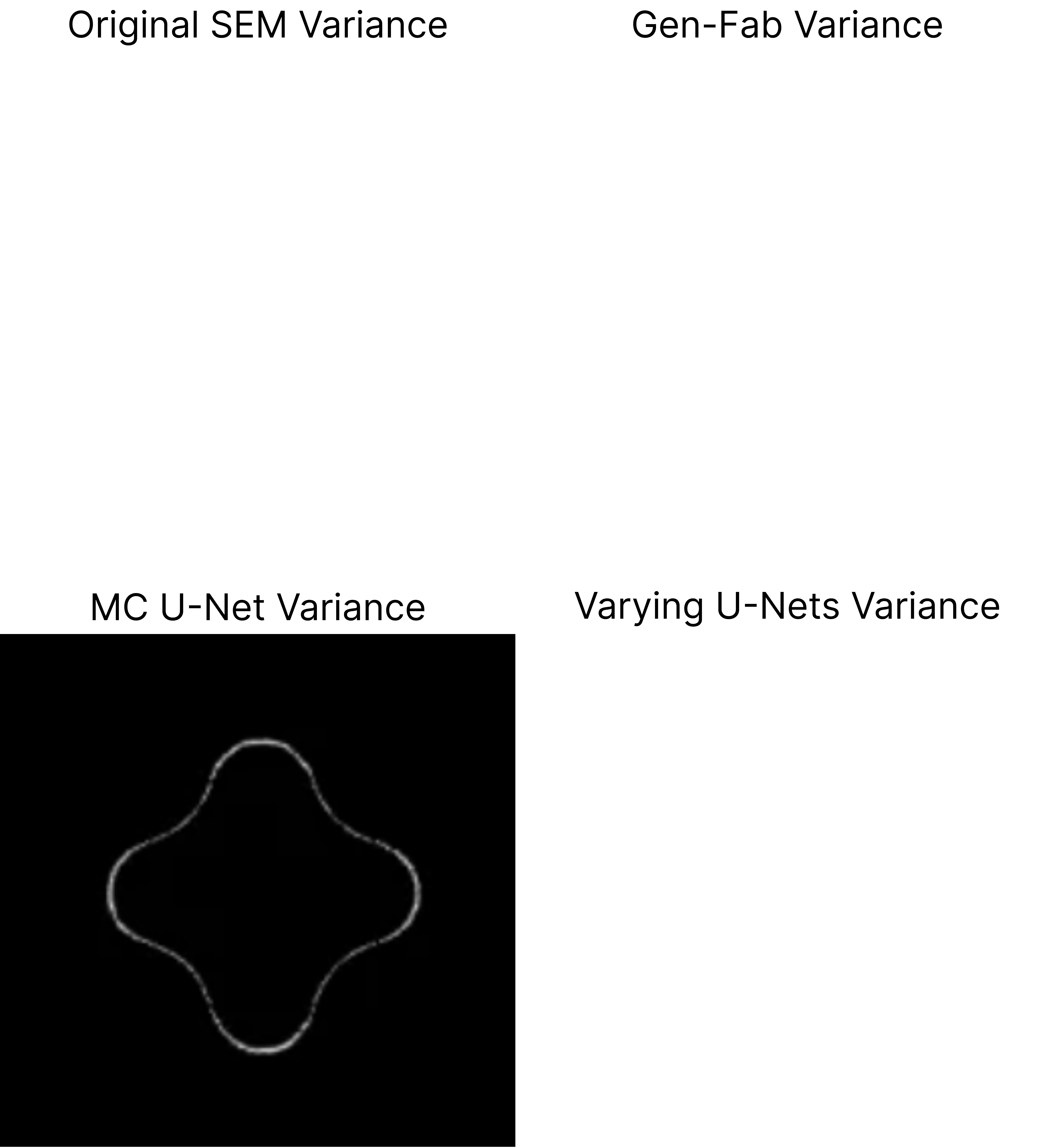}
\caption{
\textbf{Variance maps of fabricated and predicted SEM outputs for the Cross-50.}
Variance maps for Cross-50 show how Gen-Fab, MC-Dropout U-Net, and Varying U-Nets capture edge variations, calculated over 35 real and 35 generated samples per model. Gen-Fab most closely reflects real SEM variability, whereas the MC-Dropout U-Net exhibits weaker variance and and Varying U-Nets exhibit overestimated variability.}\label{fig5}
\end{figure}

\begin{figure*}[t]
\centering
\includegraphics[width=0.9\textwidth]{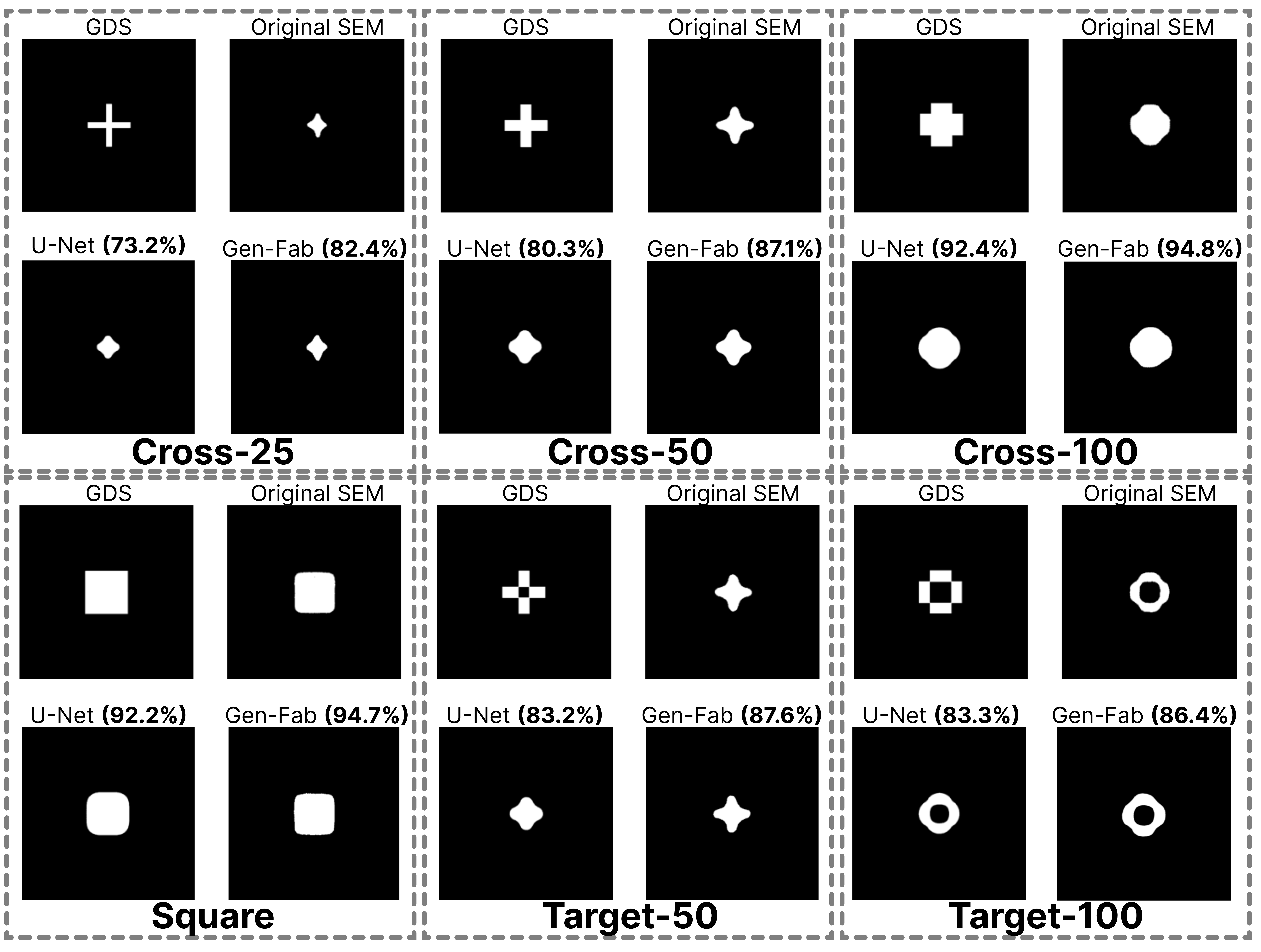}
\caption{
\textbf{Qualitative comparison of Varying U-Nets and Gen-Fab predictions across six evaluation structures.}
Each sub-panel shows the input GDS, corresponding real SEM, and predicted outputs from Varying U-Nets and Gen-Fab. IoU scores (\%) for each model are reported. Gen-Fab consistently achieves higher IoU than Varying U-Nets, indicating closer alignment with actual fabricated outcomes.}\label{fig4}
\end{figure*}

  To complement our quantitative evaluations, we qualitatively compare the outputs of each model against real SEM images to assess visual realism and structural fidelity. This analysis is particularly important given that our test geometries were intentionally chosen to differ significantly from the training dataset, both to evaluate generalization and to enable meaningful visual comparisons beyond familiar patterns. Fig.~\ref{fig4} presents side-by-side examples for each of the six test structures. For each design, we show the input GDS, the real SEM, and the predicted outputs from both the Varying U-Nets and Gen-Fab, annotated with their IoU scores based on Greedy matching.

  The U-Net predictions appear overly smooth, lacking the subtle edge roughness, asymmetry, and pattern deformations visible in the real SEM images. This is expected from a deterministic model trained to minimize pixel-wise losses, which tend to average out stochastic variations, resulting in overly symmetric predictions.

  By contrast, the Gen-Fab model produces outputs that more closely resemble real SEMs. Its predictions exhibit minor geometric irregularities, such as asymmetrical arm thickness and edge perturbations that mimic fabrication imperfections observed in every structure.

  Fig.~\ref{fig5} further highlights these distinctions by visualizing pixel-wise variance across multiple outputs from each model for the Cross-50 structure. The real SEM variance map shows variation concentrated along structure edges. The U-Net exhibits no variance across outputs, as it is deterministic by design. The MC-Dropout U-Net shows degraded generalization performance, as the smaller cross features are not well captured compared to Gen-Fab. The Varying U-Nets introduces exaggerated variations, while Gen-Fab’s variance pattern closely matches the real SEM data.

  These visualizations show that while MC-Dropout U-Net and Varying U-Nets reproduce the nominal shape, they struggle to reflect fabrication-induced randomness accurately. Gen-Fab, on the other hand, generates both structurally accurate and visually realistic SEM predictions.

\subsection{Distribution Shift Analysis}
To quantify the domain gap between the training and OOD test sets, we conducted a feature-space distribution analysis based on pretrained convolutional embeddings. Specifically, images from both domains were passed through a VGG16 feature extractor pretrained on ImageNet, and the resulting embeddings were used to characterize each distribution.

\begin{figure}[t]
    \centering
    \includegraphics[width=0.45\textwidth]{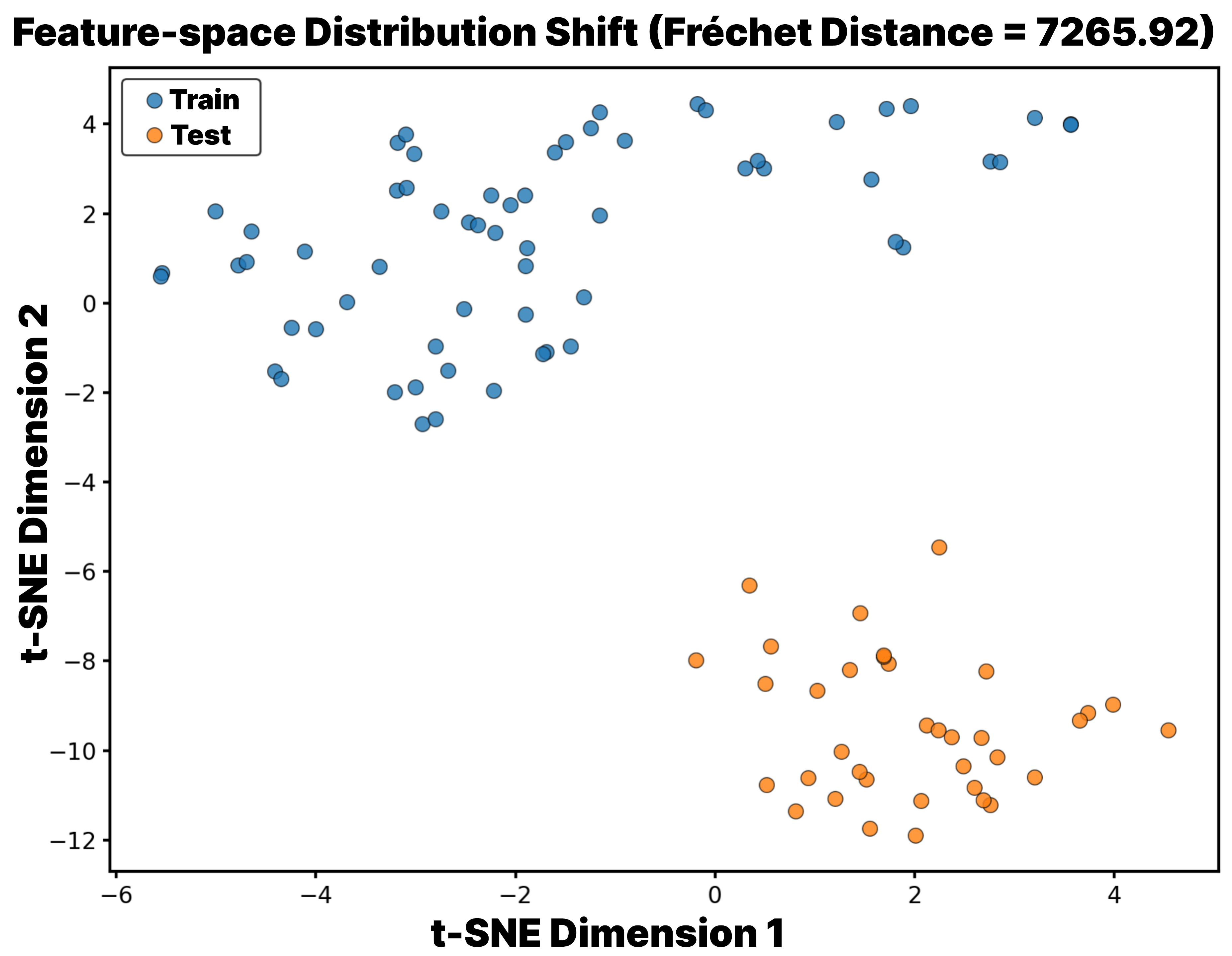}
    \caption{Feature-space distribution shift between the training and OOD test sets, visualized via t-SNE projection. The OOD test samples were intentionally selected to differ from the training distribution to evaluate the model’s generalization ability.}
    \label{fig:dist_shift}
\end{figure}

We employed two complementary metrics: (1) the Fréchet Distance (FD), which measures the divergence between the mean and covariance of the two feature distributions~\citep{heusel2017gans}, and (2) a two-dimensional t-SNE projection for qualitative visualization~\citep{vandermaaten08a}. As shown in Fig.~\ref{fig:dist_shift}, the t-SNE embedding reveals a clear separation between the training and test domains. Quantitatively, the computed FD between the training and test distributions is 7265.92. For reference, the FD between two random splits of the training set is 1912.6, and between two random splits of the test set is 1265.61, confirming that the inter-domain shift is significantly larger than intra-domain variability. Hence, the chosen test structures represent a meaningful yet realistic shift in the underlying fabrication space, suitable for evaluating the model’s generalization capability to unseen geometries.

\subsection{Epistemic vs. Aleatoric Uncertainty in Gen-Fab}

The variability observed across repeated SEM images, as shown in Fig.~\ref{fig5} (top-left), reflects aleatoric (data) uncertainty, which represents the inherent randomness of the fabrication process itself arising from stochastic lithography, etching, and imaging effects. Gen-Fab models this form of uncertainty by sampling different latent noise vectors $z$, each representing one realization of the fabrication outcome $Y$ for a given layout $X$. In contrast, epistemic (model) uncertainty originates from the imperfectly estimated model parameters $\theta$ due to limited training data. To quantify this component, we extended our uncertainty estimation using an ensemble-based approach following the law of total variance:
\[
\mathrm{Var}_{\theta,z}(Y|X) = \mathbb{E}_{\theta}\!\left[\mathrm{Var}_{z}(Y|X,\theta)\right] + \mathrm{Var}_{\theta}\!\left(\mathbb{E}_{z}[Y|X,\theta]\right),
\]
where the first term represents the expected data-dependent (aleatoric) variance across latent realizations, and the second term captures model-dependent (epistemic) variance across independently trained Gen-Fab networks. In practice, five separate Gen-Fab instances $\{G_{\theta_k}\}_{k=1}^{5}$ were trained with different random initializations and data shuffling, each evaluated using 35 latent noise vectors $z_i \sim \mathcal{N}(0,I)$. For each model, the per-pixel mean prediction $\mu_k(X)$ and intra-model variance $\sigma_k^2(X)$ were computed, and ensemble statistics were aggregated as:

\begin{equation}
\begin{aligned}
\bar{\mu}(X) &= \frac{1}{K} \sum_{k} \mu_k(X), \\
\sigma^2_{\text{aleatoric}}(X) &= \frac{1}{K} \sum_{k} \sigma_k^2(X), \\
\sigma^2_{\text{epistemic}}(X) &= \frac{1}{K}\sum_{k} \big(\mu_k(X) - \bar{\mu}(X)\big)^2.
\end{aligned}
\end{equation}

\begin{figure*}[t]
    \centering
    \includegraphics[width=1.0\textwidth]{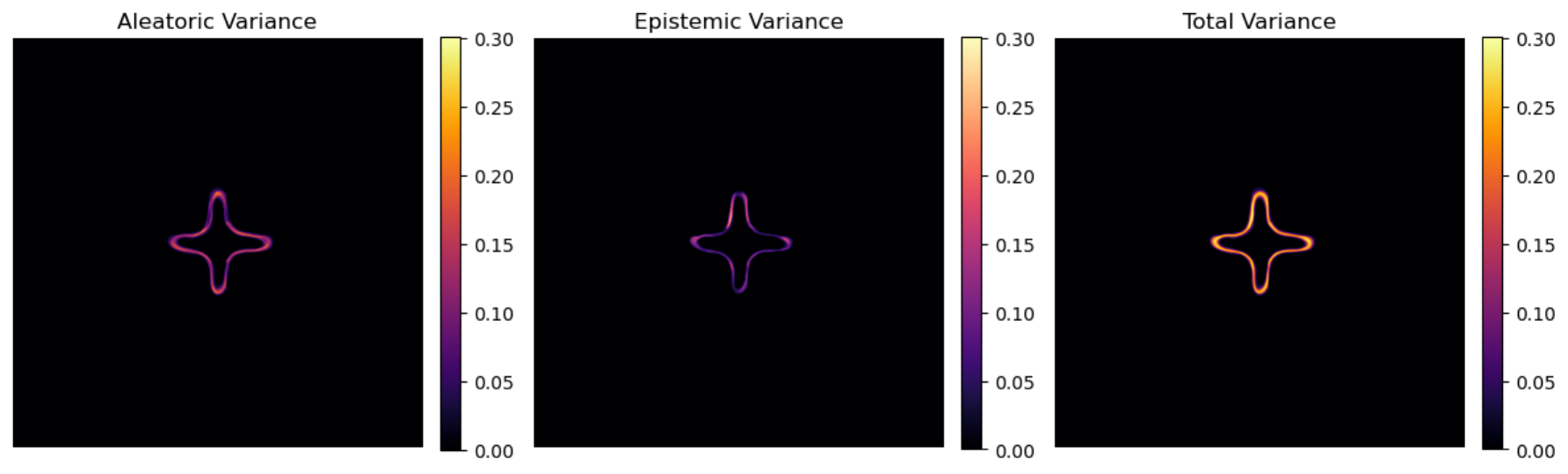}
    \caption{\textbf{Aleatoric, epistemic, and total variance decomposition for Gen-Fab.}
    Variance maps for a Cross-50 structure illustrate the uncertainty decomposition using five independently trained Gen-Fab models and 35 latent noise samples per model.
    The aleatoric variance (left) dominates along the structural edges, whereas the epistemic variance (center) remains comparatively low.
    The total variance (right) is therefore largely determined by aleatoric effects, confirming that Gen-Fab effectively captures fabrication-induced variability while maintaining low model uncertainty.}
    \label{fig:uncertainty_decomp}
\end{figure*}

As shown in Fig.~\ref{fig:uncertainty_decomp}, the aleatoric variance is concentrated along feature boundaries, consistent with real fabrication-induced edge deviations, while the epistemic variance is noticeably smaller, suggesting high inter-model consistency.
This result indicates that the primary source of predictive uncertainty arises from the stochastic fabrication process itself rather than model instability, confirming that Gen-Fab provides reliable predictions.

\subsection{Discussion and Implications}
  \subsubsection{Effectiveness of Conditional Generative Modeling}
  Gen-Fab’s advantage stems from its cGAN framework, which contrasts with deterministic models like U-Net. While U-Nets minimize pixel-wise loss to a fixed ground-truth, Gen-Fab learns a conditional distribution by introducing a latent noise vector z and training against a discriminator that enforces realism. This yields sharper and more realistic SEM predictions with diversity reflective of true fabrication variability. From a digital-twin standpoint, this distinction is critical: deterministic predictors correspond to single-state approximations of the physical process, whereas Gen-Fab enables a stochastic digital twin capable of mimicking a real fabrication process by producing multiple plausible realizations of it.

  Empirically, Gen-Fab outperforms U-Net, MC-Dropout U-Net, and Varying U-Nets across all established pixel-wise and distributional metrics. Unlike MC-Dropout U-Net and Varying U-Nets that approximate variation, Gen-Fab directly learns and samples from the fabrication process’s underlying stochasticity.
  
  \subsubsection{Accuracy vs. Variation Trade-offs}
  A key concern in generative modeling is balancing fidelity and diversity, avoiding the so-called mode collapse while producing accurate samples~\citep{srivastava2017veeganreducingmodecollapse, bau2019seeinggangenerate}. Our results show Gen-Fab achieves this trade-off effectively. Notably, even under random matching, the Gen-Fab model outperforms the U-Net baseline under greedy matching (88.7\% vs. 85.3\%), indicating that both fidelity and diversity are improved.

  If output diversity were excessive, the random-match IoU would degrade. Model hyperparameter tuning helps achieve the best balance. Moderate latent sizes (16–64) perform best, likely due to better diversity control and discriminator stability. Top models also leverage data augmentation and sufficient training duration (10k–20k steps).
  
  \subsubsection{Implications for Robust Photonic Design}
  Modeling fabrication variability has immediate applications in design-for-manufacturability and optimization. Designers can sample multiple outputs from Gen-Fab for a given GDS input, analyze critical dimension statistics, or simulate performance yield. This supports robust design strategies based on expected value, variance reduction, or percentile-based constraints.

    Within a digital-twin framework, such capabilities support virtual experimentation and Monte Carlo–style analysis prior to fabrication, allowing design robustness to be assessed directly on the twin rather than through repeated physical runs. The low KL-D and W-D (Table~\ref{tab:kl_wd_combined}) affirm that Gen-Fab’s outputs are statistically aligned with true SEM distributions, a prerequisite for reliable digital-twin deployment. This fidelity is critical for identifying edge-case failures such as waveguide narrowing or hole collapse~\citep{gostimirovic2023improving, usingml2022pattern}. Integrating such generative models into photonic CAD tools could enable automated digital-twin-driven optimization loops, where robustness is evaluated over Monte-Carlo sampled geometries rather than idealized layouts or corner cases~\citep{advancing2023ml, ma2024boson1}.

\section{Conclusion and Future Work}\label{sec5}
In this work, we presented Gen-Fab, a conditional generative model for predicting and analyzing fabrication-induced variations in nanophotonic devices. By introducing a latent noise vector into the Pix2Pix framework, our model captures the one-to-many nature of the fabrication process, generating diverse and realistic SEM predictions from a single layout design. From a digital-twin perspective, Gen-Fab enables a true fabrication-level digital twin that bridges photonic design layouts and manufactured outcomes through a fast, data-driven surrogate model. Future work will focus on close integration of GenFab with the fabrication facility to enable continuous model updates offering both up-to-date model for design purposes as well as tracking and identification of process drifts. We demonstrated that Gen-Fab significantly outperforms deterministic, MC-Dropout, and ensemble-based baselines in both accuracy and uncertainty modeling, achieving higher IoU scores and lower distributional divergence from real fabrication data. Furthermore, a distribution-shift analysis between the training and out-of-distribution (OOD) test sets confirmed that Gen-Fab maintains its predictive fidelity under realistic domain shifts, validating its generalization capability across unseen fabrication geometries.

Future work will focus on applying Gen-Fab to robust inverse design pipelines, where the model can be used to simulate a distribution of fabrication outcomes and guide the design of layouts that are not only performant in ideal conditions but also resilient to process variability. The outputs from Gen-Fab can also be directly integrated into photonic simulation tools for quantitative analysis. Because Gen-Fab produces binary SEM-like predictions aligned to the original layout, these outputs can be readily converted into polygonal geometries compatible with standard electronic design automation formatting such as GDSII, as well as electromagnetic and circuit-level simulators (e.g., Ansys Lumerical), enabling subsequent statistical geometric and optical performance evaluation under realistic process variations.

In future work, Gen-Fab could be embedded within automated inverse-design or optimization pipelines to provide fabrication-aware feedback during layout generation, closing the loop between design, simulation, and manufacturability. Extending the model to handle more complex photonic components, multilayer fabrication stacks, and broader design spaces will enhance its generalizability. Future research will focus on investigation of alternative generative frameworks such as diffusion models~\citep{ho2020denoising, rombach2022highresolution}, normalizing flows~\citep{rezende2015flows, papamakarios2021normalizing}, and style-based GANs~\citep{stylegan2021}, which may offer improved controllability, stability, and diversity for fabrication-aware design.

A further direction for future work is to integrate the interpretability into the Gen-Fab’s latent space, allowing to control distinct aspects of fabrication variability such as corner rounding or gap-filling independently.

{\small
\bmhead{Acknowledgements}
We gratefully acknowledge Applied Nanotools Inc. for their support in providing high quality SEM data used in this study.

\bmhead{Authors Contributions}
Conceptualization: RA, YG; methodology: RA, YG, implementation: RA; experiments: RA, original draft: RA;  manuscript review and editing: all authors.

\bmhead{Funding}
This work is supported by the National Research Council Canada Challenge Programs AI for Design (Grant AI4D-144).

\bmhead{Data Availability}
The data that support the findings of this study are available from the corresponding author upon reasonable request.

\section*{Declarations}

\bmhead{Conflict of Interest}
The authors declare that they have no conflict of interest.

\bmhead{Replication of Results}
The methodology described in this paper can be replicated by readers using publicly available tools. Our Gen-Fab model builds upon the standard Pix2Pix architecture, with a key modification of injecting a stochastic latent vector at the generator bottleneck described in text in detail. All architectural details, loss functions, and training protocols are described in detail throughout the paper. We also provide full descriptions of our experimental setup, evaluation metrics (IoU, KL divergence, and Wasserstein distance), and hyperparameter choices.

\bmhead{Ethics approval}
Not applicable.

\bmhead{Consent to participate}
Not applicable.
}

\begin{appendices}

\section{Convergence Analysis}\label{secA1}

This appendix presents the training curves of the proposed Gen-Fab conditional generative model in order to illustrate its convergence behavior and optimization stability.

\begin{figure}[H]
    \centering
    \includegraphics[width=0.48\textwidth]{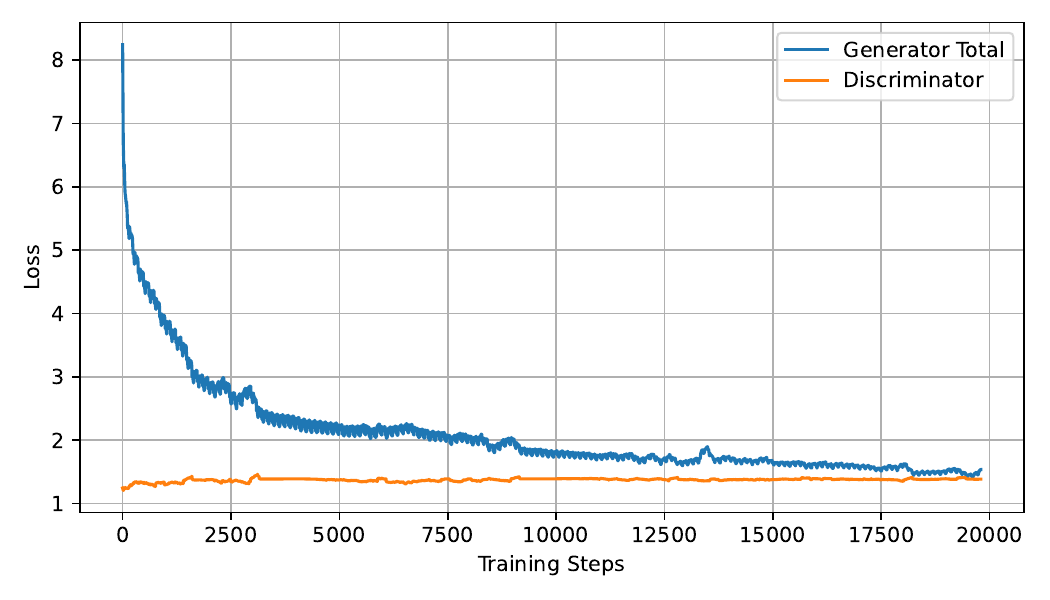}
    \caption{Generator total loss and discriminator loss during training, illustrating stable adversarial convergence of the Gen-Fab model.}
    \label{fig:gan_convergence}
\end{figure}

Figure~\ref{fig:gan_convergence} shows the evolution of the generator total loss and the discriminator loss over training steps. The generator loss decreases from a high value and gradually stabilizes, indicating consistent improvement in synthesis quality. The discriminator loss converges to a stable range without collapsing, suggesting a balanced adversarial game in which neither network dominates the training process. This behavior is characteristic of stable conditional GAN training and indicates that the model reaches a dynamic equilibrium.

\begin{figure}[H]
    \centering
    \includegraphics[width=0.48\textwidth]{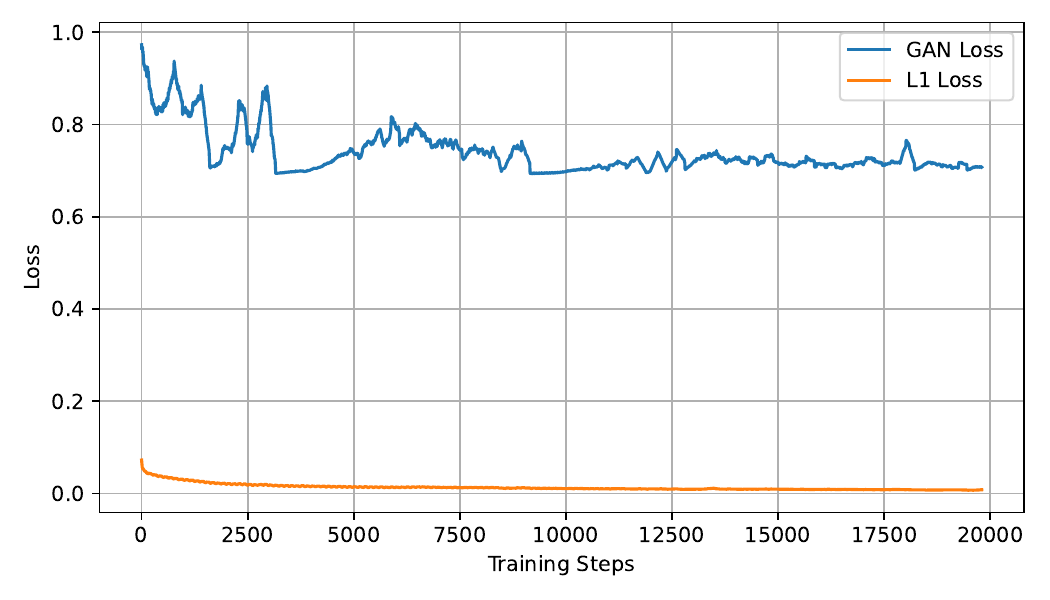}
    \caption{Evolution of the generator loss components, showing early convergence of the L1 reconstruction loss and stabilization of the adversarial (GAN) loss.}
    \label{fig:generator_components}
\end{figure}

Figure~\ref{fig:generator_components} decomposes the generator objective into its adversarial (GAN) loss and reconstruction (L1) loss components. The L1 loss rapidly decreases during early training. In contrast, the adversarial loss stabilizes at a higher value with mild oscillations, which is expected in adversarial training and indicates continued refinement of fine-scale texture realism without sacrificing structural fidelity.

\end{appendices}

\bibliography{sn-bibliography} 
\end{document}